\documentclass[11pt]{article}
\usepackage{arabtex}
\usepackage{utf8}
\usepackage{coling2020}
\usepackage{times}
\usepackage{url}
\usepackage{latexsym}
\usepackage{multirow}
\usepackage{makecell}
\usepackage{graphicx}
\usepackage{url}
\usepackage{paralist}
\usepackage{tabularx}
\usepackage{qtree}
\usepackage{arydshln}
\usepackage{wrapfig,lipsum,booktabs}
\usepackage{hyperref}
\usepackage{booktabs}
\usepackage{colortbl}
\usepackage{color,soul}
\hypersetup{
    colorlinks=true,
    linkcolor=blue,
    filecolor=blue,  
    citecolor=blue,
    urlcolor=blue,
}
\usepackage{comment}

\usepackage{adjustbox}
\usepackage{color, colortbl}
\usepackage[table]{xcolor}
\usepackage[toc,page]{appendix}
\usepackage{tikz}
\def\checkmark{\tikz\fill[scale=0.4](0,.35) -- (.25,0) -- (1,.7) -- (.25,.15) -- cycle;} 

\definecolor{light-gray}{gray}{0.85}

\usepackage{footnote}
\makesavenoteenv{tabular}
\makesavenoteenv{table}

\title{NADI 2020: The First Nuanced Arabic Dialect Identification Shared Task}

\author{Authors}

\author{Muhammad Abdul-Mageed, Chiyu Zhang, Houda Bouamor,$^\dagger$ Nizar Habash$^\ddagger$\\
The University of British Columbia, Vancouver, Canada\\
$^\dagger$Carnegie Mellon University in Qatar, Qatar\\
$^\ddagger$New York University Abu Dhabi, UAE\\
  {\tt muhammad.mageed@ubc.ca ~~~~ chiyuzh@mail.ubc.ca}\\
  {\tt ~~ hbouamor@cmu.edu ~~~~ nizar.habash@nyu.edu}\\
  }
\date{}

\begin{document}

\setarab 
\maketitle
\setcode{utf8}

\begin{abstract}
We present the results and findings of the First Nuanced Arabic Dialect Identification Shared Task (NADI). 
This Shared Task includes two subtasks: country-level dialect identification (Subtask 1) and province-level sub-dialect identification (Subtask 2). 
The data for the shared task covers a total of 100 provinces from 21 Arab countries and are collected from the Twitter domain.  As such, NADI is the first shared task to target naturally-occurring fine-grained dialectal text at the sub-country level. 
A total of 61 teams from 25 countries registered to participate in the tasks, thus reflecting the interest of the community in this area. We received 47 submissions for Subtask 1 from 18 teams and 9 submissions for Subtask 2 from 9 teams. 

\end{abstract}

\section{Introduction}\label{sec:intro}

\blfootnote{
    
     \hspace{-0.65cm}  
     This work is licensed under a Creative Commons 
     Attribution 4.0 International License.
     License details:
     \url{http://creativecommons.org/licenses/by/4.0/}.
}

\begin{wrapfigure}{r}{0.5\textwidth}
  \begin{center}
  \frame{\includegraphics[width=8cm,height=4.5cm]{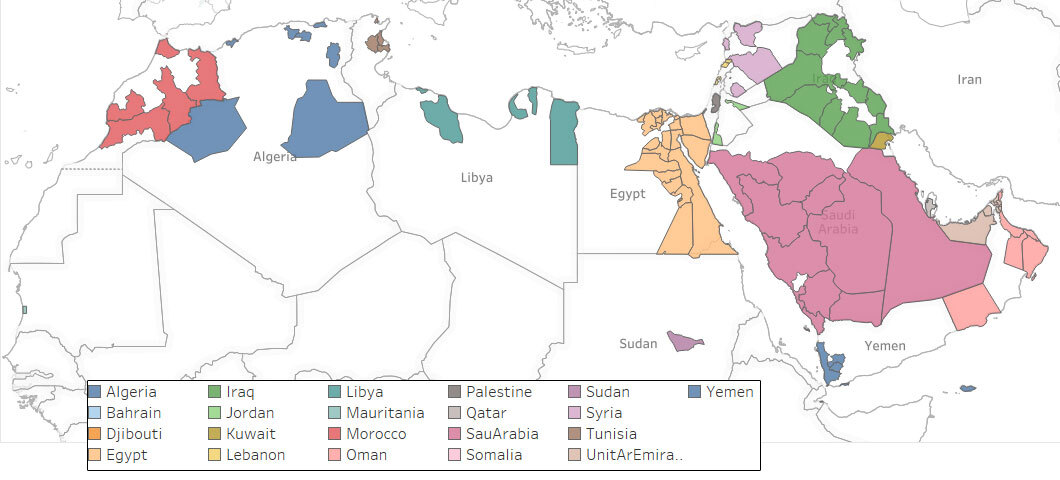}}
  \end{center}
\caption{A map of the Arab World showing countries and provinces in the NADI dataset. Each of the 21 countries is represented in a color different from that of a neighboring country. Provinces are marked with lines inside each country.}
\label{fig:province_map}
\end{wrapfigure}

The Arab world is an extensive geographical region across Africa and Asia, with a population of $\sim 400$ million people whose native tongue is Arabic. Arabic could be classified into three major types: (1) Classical Arabic (CA), the language of the Qur'an and early literature, (2) Modern Standard Arabic (MSA), the medium used in education and formal and pan-Arab media, and (3) dialectal Arabic (DA), a host of geographically and politically defined variants. Modern day Arabic is also usually described as a \textit{diglossic} language with a so-called `High' variety that is used in formal settings (MSA), and a `Low' variety that is the medium of everyday communication (DA). The presumably  `Low variety' is in reality a collection of variants. One axis of variation for Arabic is geography where people from various sub-regions, countries, or even provinces within the same country, may be using language differently. 

The goal of the First Nuanced Arabic Dialect Identification (NADI) Shared Task is to provide resources and encourage efforts to investigate questions focused on dialectal variation within the collection of Arabic variants. The NADI shared task targets 21 Arab countries and a total of 100 provinces across these countries. The shared task consists of two subtasks: \textit{country-level} dialect identification (Subtask~1) and  \textit{province-level} detection (Subtask~2). We provide participants with a new Twitter labeled dataset that we collected exclusively for the purpose of the shared task. 
The dataset is publicly available for research.\footnote{The dataset is accessible at the shared task page: \url{http://nadi2020.arabic-nlp.net}.} A total of 52 teams registered for the shard task, of whom 18 teams ended up submitting their systems for scoring. We then received 15 papers, of which we accepted 14.

This paper is organized as follows.  We provide a brief overview of the computational linguistic literature on Arabic dialects in Section~\ref{sec:rel}. 
We describe the two subtasks and dataset in Sections~\ref{sec:task} and Section~\ref{sec:data}, respectively. And finally, we introduce participating teams, shared task results, and a high-level description of submitted systems in Section~\ref{sec:teams_res}. 
\section{Related Work}\label{sec:rel}
 As we explained in Section~\ref{sec:intro}, Arabic could be viewed as comprised of 3 main types: CA, MSA, and DA. While CA and MSA have been studied and taught extensively, DA has only received more attention relatively recently \cite{Harrell:1962:short,Cowell:1964:reference,badawi1973levels,Brustad:2000:syntax,Holes:2004:modern}.

A majority of DA computational efforts have targeted creating resources for country or regionally specific dialects~\cite{Gadalla:1997:callhome,diab2010colaba,al2012yadac,sadat2014automatic,Smaili:2014:building,Jarrar:2016:curras,Khalifa:2016:large,Al-Twairesh:2018:suar,el-haj-2020-habibi}. 
The expansion into multi-dialectal data sets and models to identify them was initially done at the regional level 
\cite{zaidan2011arabic,Elfardy:2014:aida,Bouamor:2014:multidialectal,Meftouh:2015:machine}. A number of Arabic dialect identification shared tasks were organized as part of the VarDial workshop. These focused on regional varieties such as Egyptian, Gulf, Levantine, and North African based on speech broadcast transcriptions~\cite{malmasi2016discriminating} but also acoustic features \cite{zampieri2017findings} and phonetic features~\cite{zampieri2018language} extracted from raw audio.  \newcite{althobaiti2020automatic} presents a recent survey of computational work on Arabic dialects.

An early effort for creating finer grained parallel dialectal corpus and lexicon was done under the Multi Arabic Dialects Application and Resources (MADAR) project ~\cite{Bouamor:2018:madar}. 
The parallel data was created by commission under controlled settings to maximize its use for cross-dialectal comparisons and machine translation. 
Their data was also used for dialectal identification at the city level \cite{Salameh:2018:fine-grained,obeid-etal-2019-adida} of 25 Arab cities.  One issue with the MADAR data in the context of identification is that it was commissioned and not naturally occurring.  Concurrently, larger Twitter-based datasets covering 10-21 countries were also introduced~\cite{Mubarak:2014:using,Abdul-Mageed:2018:you,Zaghouani:2018:araptweet}. Researchers are also starting to introduce DA datasets labeled for socio-pragmatics, e.g., ~\cite{abbes2020daict,mubarak2020overview}. The MADAR shared task~\cite{bouamor2019madar} comprised two subtasks, one focusing on 21 Arab countries exploiting Twitter data manually labeled at the user level, and another on 25 Arab cities mentioned above. During the same time as NADI,~\newcite{mageed2020microdialect} describe data and models at country, province, and city levels.

The NADI shared task follows these pioneering works by availing data to the (Arabic) NLP community, and encouraging work on Arabic dialects. Similar to the MADAR shared task, we include a country-level dialect identification task (Subtask~1), and a sub-country dialect identification task (Subtask~2).
However, our sub-country task is a province-level identification task with a much larger label set than MADAR's city-level task, and is based on naturally occurring data. We hope that our work will be setting the stage for exploring variation in geographical regions that have not been studied before.
\section{Task Description}\label{sec:task}
The NADI shared task consists of two subtasks for country-level and province-level classification.

\subsection{Subtask~1: Country-level Classification}

The goal of Subtask~1 is to identify country-level dialects from short written sentences (tweets). NADI Subtask~1 is similar to previous works that have also taken country as their target  \cite{Mubarak:2014:using,Abdul-Mageed:2018:you,Zaghouani:2018:araptweet,bouamor2019madar}.
Labeled data was provided to NADI participants with specific TRAIN and development (DEV) splits. Each of the 21 labels corresponding to the 21 countries is represented in both TRAIN and DEV. Teams could score their models through an online system on the DEV set before the deadline. Our TEST set of unlabeled tweets was released shortly before the system submission deadline. Participants were invited to submit their predictions to the online scoring system that housed the gold TEST set labels. We provide the distribution of the TRAIN, DEV, and TEST splits across countries in Table~\ref{tab:country}. 

\begin{table}[t]
\centering
\footnotesize
\begin{tabular}{l|r|r|r|r|r|r}
\hline
\multicolumn{1}{c|}{\multirow{2}{*}{\textbf{\begin{tabular}[c]{@{}c@{}}Country \\ Name\end{tabular}}}} & \multicolumn{1}{c|}{\multirow{2}{*}{\textbf{\begin{tabular}[c]{@{}c@{}}\# of \\ Provinces\end{tabular}}}} & \multicolumn{5}{c}{\textbf{\# of Tweets}}                                                                                                                                            \\ \cline{3-7} 
\multicolumn{1}{c|}{}                                                                                  & \multicolumn{1}{c|}{}                                                                                   & \multicolumn{1}{c|}{\textbf{Train}} & \multicolumn{1}{c|}{\textbf{Dev}} & \multicolumn{1}{c|}{\textbf{Test}} & \multicolumn{1}{c|}{\textbf{Total}} & \multicolumn{1}{c}{\textbf{\%}} \\ \hline
Algeria                                                                                                 & 7                                                                                                       & 1,491                               & 359                               & 364                                & 2,214                               & 7.15                            \\ 
Bahrain                                                                                                 & 1                                                                                                       & 210                                 & 8                                 & 20                                 & 238                                 & 0.77                            \\ 
Djibouti                                                                                                & 1                                                                                                       & 210                                 & 10                                & 51                                 & 271                                 & 0.88                            \\ 
Egypt                                                                                                   & 21                                                                                                      & 4,473                               & 1,070                             & 1,092                              & 6,635                               & 21.43                           \\ 
Iraq                                                                                                    & 12                                                                                                      & 2,556                               & 636                               & 624                                & 3,816                               & 12.33                           \\ 
Jordan                                                                                                  & 2                                                                                                       & 426                                 & 104                               & 104                                & 634                                 & 2.05                            \\ 
Kuwait                                                                                                  & 2                                                                                                       & 420                                 & 70                                & 102                                & 592                                 & 1.91                            \\ 
Lebanon                                                                                                 & 3                                                                                                       & 639                                 & 110                               & 156                                & 905                                 & 2.92                            \\ 
Libya                                                                                                   & 5                                                                                                       & 1,070                               & 265                               & 265                                & 1,600                               & 5.17                            \\ 
Mauritania                                                                                              & 1                                                                                                       & 210                                 & 40                                & 5                                  & 255                                 & 0.82                            \\ 
Morocco                                                                                                 & 5                                                                                                       & 1,070                               & 249                               & 260                                & 1,579                               & 5.10                            \\ 
Oman                                                                                                    & 6                                                                                                       & 1,098                               & 249                               & 268                                & 1,615                               & 5.22                            \\ 
Palestine                                                                                               & 2                                                                                                       & 420                                 & 102                               & 102                                & 624                                 & 2.02                            \\ 
Qatar                                                                                                   & 2                                                                                                       & 234                                 & 104                               & 61                                 & 399                                 & 1.29                            \\ 
Saudi Arabia                                                                                            & 10                                                                                                      & 2,312                               & 579                               & 564                                & 3,455                               & 11.16                           \\ 
Somalia                                                                                                 & 1                                                                                                       & 210                                 & 51                                & 51                                 & 312                                 & 1.01                            \\ 
Sudan                                                                                                   & 1                                                                                                       & 210                                 & 51                                & 51                                 & 312                                 & 1.01                            \\ 
Syria                                                                                                   & 5                                                                                                       & 1,070                               & 265                               & 260                                & 1,595                               & 5.15                            \\ 
Tunisia                                                                                                 & 4                                                                                                       & 750                                 & 164                               & 208                                & 1,122                               & 3.62                            \\ 
UAE                                                                                                     & 5                                                                                                       & 1,070                               & 265                               & 213                                & 1,548                               & 5.00                            \\ 
Yemen                                                                                                   & 4                                                                                                       & 851                                 & 206                               & 179                                & 1,236                               & 3.99                            \\ 
\textbf{Total}                                                                                                   & \textbf{100 }                                                                                                    & \textbf{21,000}                             & \textbf{4,957}                          & \textbf{5,000}                              & \textbf{30,957}                              & \textbf{100.00}                          \\ \hline
\end{tabular}
\caption{Distribution of country-level dialect identification data for Subtask~1 across our data splits. }\label{tab:country}
\end{table}

\subsection{Subtask~2: Province-level Classification}
The goal of Subtask~2 is to identify the specific state or province (henceforth, {\it province}) from a list of 100 provinces. The provinces are unequally distributed among the list of 21 countries. While efforts on city-level and country-level prediction were the topic of a previous shared task \cite{bouamor2019madar}, to the best of our knowledge, the target of automatic dialect prediction  at a small geographical region such as a province has not been previously investigated, thus lending novelty to this subtask. We acknowledge that this subtask has some affinity to work focused on predicting geolocation based on tweets. Nevertheless, geolocation prediction is performed at the level of users not tweets and hence is different. There are also differences between our work here and geolocation as to how the data was collected. We further explain this nuance in Section~\ref{sec:data}. The distribution of the classes across the 100 provinces in our data splits is presented in Table~\ref{tab:data_provinces} in Appendix A. 

For both Subtask~1 and Subtask~2, tweets in the TRAIN, DEV and TEST splits come from distinct sets of \textit{users}, such that no user had their tweets in any two of the TRAIN, DEV, and TEST splits.

\subsection{Restrictions and Evaluation Metrics}
To ensure fair comparisons and common experimental conditions, we provided participating teams with a set of restrictions that apply to the two subtasks, and clear evaluation metrics. Our method of distributing the data as well as our evaluation setup through the CodaLab online platform also facilitated the competition management, enhanced timeliness of acquiring results upon system submission, and guaranteed ultimate transparency.\footnote{\url{https://codalab.org/}}

We directly provided participants with the actual tweets posted to the Twitter platform, rather than tweet IDs. This enabled comparison between systems exploiting identical data. Since we shared actual tweets, we did not share tweet IDs with participants. This made it harder to collect data from the same user from which a tweet comes. For the two subtasks, we asked to only and exclusively use our distributed data. In other words, we provided instructions not to use any external data nor search or depend on any additional user-level information such as geolocation. In addition to our labeled TRAIN and DEV splits, we provided tweet IDs for 10M tweets and a simple script that can be used to collect the tweets. We did not provide any labels for this additional 10M tweet set, and encouraged participants to use it in developing their models in any way they deemed useful.

For both subtasks, the official metric is macro-averaged $F{_1}$ score obtained on blind test sets. We also report performance in terms of macro-averaged precision, macro-averaged recall and accuracy for systems submitted to each of the two subtasks. Each participating team was allowed to submit up to five runs for each subtask, and only the highest scoring run was kept as representing the team. Although official results are based only on a blind TEST set, we also asked participants to report their results on the DEV set in their papers. We setup two CodaLab competitions for scoring participant systems.\footnote{The 
CodaLab competition for Subtask~1 is accessible at: \url{https://competitions.codalab.org/competitions/24001}.}$^{,}$\footnote{The 
CodaLab competition for Subtask~2 is accessible at: \url{https://competitions.codalab.org/competitions/24002}.} We will keep the Codalab competition for each task live post competition, for researchers who would be interested in training models and evaluating their systems using the shared task TEST set.


\begin{table}[t]

\centering 
\setlength{\tabcolsep}{2pt}

\footnotesize
\begin{tabular}{l|l|r}
\hline
\textbf{Country} & \textbf{Province} & \multicolumn{1}{c}{\textbf{Tweet}} \\\hline

\textbf{Algeria} & \textbf{Bordj-Bou-} & 
<وانتم سيد الفاظل ملائكة اكبر مصائبنا منكم والعربان دمرتوا كل شيء جميل > \\
&\textbf{Arreridj}& <  الله المستعان> \\\cline{2-3}

&    \textbf{Jijel} &
<ابراهيم غدوة نتهلا فيه مليح>\\\hline

\textbf{Egypt} & \textbf{Asyut} & <يا اقرع انت واخوك إبراهيم اللي زرع شعره واكيد الكل عارف جاب الشعر منين في  > \\
&& < جسمه وحطه في رأسه انت نسيت نفسك ولا ايه الزمالك هو اللي لمك من الشارع> \\
&& <بعد ما صالح سليم طردك زي الكلب نسيت وانت مدرب ماحدش طلعك السما > \\
&& < غير الزمالك ورجل شيكابالا الزمالك سيدك يا اقرع \#ادعم\_باسم\_مرسي>\\ \cline{2-3}
& \textbf{Suez} &
<يبقي حد يوريني نفسه بقي انا قولت حلووو غلط>\\\hline

\textbf{KSA} & \textbf{Ar-Riyad } & 
 <فيه كثير أمور عني ما تعرفونها ياليت ان الأمر بهالسهولة وياليتني مرتاحه فعلا > \\
&& <  ومستقله بشكل كامل ما كان لقيتيني اشتكي ولا اقول هالكلام راح ابعد عنهم كثيير > \\
&& <  ان شاء الله واهاجر وما ارجع لهالبقعة الجغرافية ابد واترك كل شي شين وراي > \\
&& <  وابدأ حياة حقيقية من أول وجديد الله يعدي الأيام بسرعه بس> 
\\\cline{2-3}
&  \textbf{Najran}   & 
<\#يسعد\_مساكم حمودي يسمي عليكم ويقول رايح يشتري ثوب \#العيد  > \\
&& < \#عيدكم\_مبارك\_وعساكم\_من\_عواده >\\\hline

\textbf{Morocco} & \textbf{Marrakech-} & 
<السيادة في الدنيا والسعادة في العقبى لا يوصل إليها إلا على جسر من المتاعب.  >\\
&\textbf{Tensift-Al-Haouz}&  <\#ابن\_القيم>\\\cline{2-3}
& \textbf{Tanger-Tetouan} &
<\#اليابان\_كولومبيا المبارة مانجا ولا فيلر>   \\\hline

\textbf{Oman} &\textbf{Muscat} & 
<بصراحة... غريبين هدول الثورجية يلي بيجلسوا ينظروا من برا وبخبروك انو هنن  > \\
&& < تركوا شغلهم واموالهم وبيتهم وهلأ عايشين برات البلاد بسبب الظلم على  > \\
&& < قولهم... طيب بالأول ما حدا طلب منك تترك البلاد... ثانيا... ليش عم تحرض الناس > \\
&& <  يلي موجود بالبلد وهي يلي لازم تتبهدل وتعتقل وحضرتك برا البلاد.>  \\\cline{2-3}
&    \textbf{Ad-Dakhiliyah}  &
<الحين بتبدي الشماته>\\\hline

\textbf{Sudan} & \textbf{Khartoum}  &
<نحن نبقى عليها والله شنطة زاتو ما دايرة ارح امفكو ساي>\\


\hline 
\textbf{UAE} & \textbf{Abu-Dhabi} & <ولو بالغلط يعني . \#اليوم\_العالمي\_للرسائل> \\ \cline{2-3}
 & \textbf{Ras-Al-Khaymah}  & <عليك أغار و أكتم هالشعور أخاف غيرتي بالحيل تزعجك.>\\ \hline

\end{tabular}
\label{tab:twt_exmpls}
\caption{Randomly picked examples from select provinces and corresponding countries.}
\end{table}

\section{Shared Task Datasets}\label{sec:data}
We distributed a single dataset with two sets of labels, one for Subtask~1 and another for Subtask~2. In other words, the same tweet occurs in each of the two subtasks but with different subtask-specific labels.
Additionally, we made available an unlabeled dataset for optional use in any of the two subtasks. We now provide more details about both the labeled and unlabeled data.  

\subsection{Data Collection}

We used the Twitter API to crawl data from 100 provinces belonging to 21 Arab countries for 10 months (Jan. to Oct., 2019).\footnote{Although we tried, we could not collect data from Comoros to cover all 22 Arab countries.} Next, we identified users who consistently and \textit{exclusively} tweeted from a single province during the whole 10 month period. We crawled up to 3,200 tweets from each of these users. 

\subsection{Data Sets}

\paragraph{Subtask~1 and Subtask~2 Data} We labeled tweets from each user with the country and province from which the user posted for the whole of the 10 months period, thus exploiting user consistent posting {\it location} as a proxy for {\it  dialect labels}. Note that this labeling method can still have issues as we explain in Section~\ref{subsec:data_issues}. We randomly sampled 30,957 tweets of length 5 words or more from the collection and split them into TRAIN (n=21,000), DEV (n=4,957), and TEST (n=5,000). Although the task is at the tweet level, we sampled the data for each of the TRAIN, DEV, and TEST from a unique set of users (i.e., users are not shared across the 3 splits). We distribute data for the two subtasks directly to participants in the form of actual tweet text (i.e., hydrated content). Tables~\ref{tab:country} and~\ref{tab:data_provinces} show the distribution of tweets across the data splits for both Subtask~1 and Subtask~2, respectively.

\begin{wrapfigure}{r}{0.5\textwidth}
  \begin{center}
  \frame{\includegraphics[width=7cm,height=4.5cm]{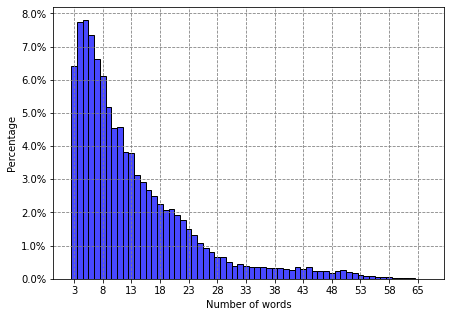}}
  \end{center}
\caption{Distribution of tweet length in words in NADI labeled data.}
\label{fig:province_map}
\end{wrapfigure}

\paragraph{Unlabeled 10M} ~We also crawled 10 million posts from Arabic Twitter during 2019. We call this dataset UNLABELED 10M and distribute it in the form of tweet IDs along with a script that can be used to crawl the actual tweets. We put no restrictions on using UNLABELED 10M for system development for either of the two subtasks.\footnote{Subtask~1, Subtask~2, and UNLABELED 10M data is available at \url{http://nadi2020.arabic-nlp.net}. More information about the data format can be found in the accompanying README file.} Next, we discuss a number of nuances and issues found in the data.

\subsection{Data Issues}\label{subsec:data_issues}

\paragraph{Location as proxy for dialect.} Our method of using consistent location (i.e., posting from the same location for at least 10 months) as a proxy for assigning dialect labels is useful, but not ideal. Even though this method allows us to collect provably relevant data, as manually verified in a small random sample of users (n=30), it can be error prone since a user with a dialect of one country can be posting from a different country during this whole period of 10 months.  

\paragraph{MSA vs. Dialect.} As we explained in Section~\ref{sec:intro}, Arabic is usually characterized as a diaglossic language with MSA being the `High' variety and DA as the `Low' variety. Arabic users also switch between these two varieties. Most relevant to our work, communication in DA over social media is not devoid of MSA even at a level as short as that of a tweet. This can vary from one dialect to another, but also depending on a range of other factors including the user educational background, career, and the actual goal of the post itself. To illustrate, based on our intuition and occasional observation, users with training in language sciences (education), those in careers such as media or higher education (job), or those trying to reach out to especially older generations or project religious or cultural authority (goal or pragmatic function) will likely use more MSA. Due to the co-existence of MSA and DA in the same tweet, we opted for using all the data we collected from the users for the competition. Alternatively, we could have identified MSA tweets either manually or automatically and removed these. We did not take that step in order to keep the task more challenging since a classifier would need to learn about patterns of MSA-DA mixing to perform well. A model would also need to acquire skills enabling it to tease apart tweets that may be overly or exclusively MSA. To explore the extent of MSA in the dataset, we run an in-house neural MSA-DA model ($acc=89.1\%, F_1=88.6\%$) on it. The model predicts the percentages of MSA data as follows: 49.5\% for TRAIN, 46.6\%for DEV, and 49.7\% for TEST. This distribution needs to be couched with caution, however, since dialectal features can be subtle and mixed with MSA to varying degrees. Upon manual inspection of a random sample of tweets the model labeled as MSA, we identify examples that are fully and verifiably MSA such as \#1 below. In addition, we observe sequences that carry dialectal features. For example \#2 and \#3 below have dialectal words (highlighted in \colorbox{orange!15}{orange}):

\begin{enumerate}

\small
    \item <الحمد لله رب العالمين على نعمه التي لا تعد ولا تحصى ولم يغفل عن رزق كل خلقه.> 
    


    \item <على بعضه يا صاحبي . قاتلهم الله> \colorbox{orange!15}{\<يتدردب>} <البيض الفاسد> 
    \item <عليهم> \colorbox{orange!15}{<تبديني>} <انت>  \colorbox{orange!15}{<أبيك>} <اخذك منهم ،>  \colorbox{orange!15}{<أبي>} <ما>

\end{enumerate}

We also observe that the more confident the MSA-DA model is (based on softmax value), the more likely its decision is correct. This suggests we can use a thresholding approach to filter out MSA tweets, should we desire to reduce MSA in the data. We leave further investigation of this issue to the future. 

\paragraph{Non-Arabic Text.} Despite efforts to exclusively keep Arabic content, our dataset had a small percentage (2.52\%) of Farsi. While collecting the data, we only kept tweets assigned an Arabic language tag by the Twitter API. However, the API is error prone and hence some non-Arabic was not filtered out. To circumvent this, we only kept tweets that have at least three word written in Arabic script after running an internal normalizer that removes diacritics and reduced repetitions of consecutive characters of $>2$ to only 2, replaced URLs and usernames with the generic strings $URL$ and $@USR$. Even after this step, some Farsi leaked to our data. The reason is that Farsi is written in the same script as Arabic, with only a few differences. \newcite{aliwy:2020:arabic-dialects} manually inspected the NADI TRAIN set and provided a distribution of Farsi tweets over the different countries. We share this distribution in Table~\ref{tab:farsi}.

\begin{wraptable}{r}{0.5\textwidth}
\centering
\footnotesize
\begin{tabular}{lrrr}
\hline
\multicolumn{1}{c}{\textbf{Country}} & \multicolumn{1}{c}{\textbf{\# tweet}} & \multicolumn{1}{c}{\textbf{\#  Farsi}} & \multicolumn{1}{c}{\textbf{\% Farsi}} \\ \hline
\textbf{Algeria}                     & 1,491                                              & 5                                       & 0.34                                       \\
\textbf{Egypt}                       & 4,473                                              & 2                                       & 0.04                                       \\
\textbf{Iraq}                        & 2,556                                              & 382                                     & 14.95                                      \\
\textbf{Morocco}                     & 1,070                                              & 1                                       & 0.09                                       \\
\textbf{Oman}                        & 1,098                                              & 26                                      & 2.37                                       \\
\textbf{Saudi Arabia}               & 2,312                                              & 6                                       & 0.26                                       \\
\textbf{Syria}                       & 1,070                                              & 3                                       & 0.28                                       \\
\textbf{Tunisia}                     & 750                                               & 2                                       & 0.27                                       \\
\textbf{UAE}      & 1,070                                              & 7                                       & 0.65                                       \\
\textbf{Yemen}                       & 851                                               & 70                                      & 8.23                                       \\ \hline
\textbf{Total}                     & 21,000                                             & 504                                     & 2.40                                       \\ \hline
\end{tabular}
\caption{Distribution of tweets manually labeled as Farsi in TRAIN. We only provide countries with Farsi tweets, and remove the rest of 21 countries in NADI.}\label{tab:farsi}
\end{wraptable}
\section{Shared Task Teams \& Results}\label{sec:teams_res}

\subsection{Our Baseline Systems}
We have two baseline classifier, Baseline I and Baseline II. \textbf{Baseline~I} is based on the majority class in the TRAIN data for each subtask. It scores at \textit{$accuracy=21.84\%$} and \textit{$F_1=1.71\%$} for Subtask 1 and \textit{$accuracy=1.92\%$} and \textit{$F_1=0.04\%$}  for Subtask 2. For \textbf{Baseline~II}, we initially train two classifiers for these two sub-tasks individually. For each task, we fine-tune on Google's pre-trained multi-lingual BERT-Base (mBERT).\footnote{https://github.com/google-research/bert} We set the maximum length of sequences in our model to 50 tokens, and employ batch training with a batch size of 8 for this model. We run the network for 20 epochs and save the model at the end of each epoch, choosing the model that performs highest on DEV as our best model. For country-level identification (Subtask 1), our best result is acquired with 16 epochs. Our best result is obtained with 20 epochs on province-level task (Subtask 2). Our mBERT model obtains \textit{$accuracy=32.38\%$} and \textit{$F_1=13.32\%$} on country-level classification and \textit{$accuracy=3.32\%$} and \textit{$F_1=2.13\%$} for province-level classification. 

\subsection{Participating Teams}
We received a total of 61 unique team registrations, among which 7 teams registered to participate in Subtask 1 only, 1 team registered to participate in Subtask 2 only, and 53 teams registered to participate in both subtasks. After evaluation phase, we received 47 submissions for Subtask 1 from 18 teams and 9 submissions for subtask 2 from 4 teams. Of participating teams, a total of 15 teams submitted description papers all of which except one were accepted for publication. Table~\ref{tab:teams} lists the 14 teams whose papers were accepted.

\begin{table}[t]
\centering
\footnotesize
\begin{tabular}{l|l|r}
\hline
\multicolumn{1}{c|}{\textbf{Team}} & \multicolumn{1}{c|}{\textbf{Affiliation}}                & \multicolumn{1}{c}{\textbf{Tasks}} \\ \hline
\textbf{Mawdoo3 AI}~\cite{talafha:2020:multi-dialect}                 & Mawdoo3 AI, Jordan                                               & 1                                   \\
\textbf{BERT\_NGRAMS}~\cite{mekki:2020:weighted}               & Mohammed VI Polytechnic University, Morocco                       & 1,2                                 \\ 
\textbf{ArabicProcessors}~\cite{gaanoun:2020:arabicprocessors}           & Institut National de Statistique et d'Economie, Morocco & 1,2                                 \\ 
\textbf{Tri-directional}~\cite{beltagy:2020:arabic-dialect}            & Faculty of Engineering, Alexandria University, Egypt            & 1                                   \\ 
\textbf{MMZ}~\cite{mansour:2020:arabic-dialect}                        & Faculty of Engineering, Alexandria University, Egypt            & 1                                   \\ 
\textbf{QMUL Team}~\cite{aloraini:2020:arabic-dialect}                  & Queen Mary University of London, United Kingdom                          & 1                                   \\ 
\textbf{Code Lyoko}~\cite{tahssin:2020:identifying}                 & Faculty of Engineering, Alexandria University, Egypt             & 1                                   \\ 
\textbf{TRY\_NLP}~\cite{balaji:2020:semi-supervised}                   & SSN College of Engineering, India                               & 1,2                                 \\ 
\textbf{Sorbonne}~\cite{ghoul:2020:comparison}                   & Sorbonne Université, France                                       & 1                                   \\ 

\textbf{Speech Translation}~\cite{lichouri:2020:speechtrans}         & CRSTDLA Research Center, Algeria                                  & 1                                   \\ 
\textbf{LTG-ST}~\cite{touileb:2020:ltg-st}                     & University of Oslo, Norway                                        & 1                                   \\ 
\textbf{Alexa}~\cite{younes:2020:team-alexa}                      & Jordan University of Science and Technology, Jordan              & 1                                   \\ 
\textbf{Alpha}~\cite{alShenaifi:2020:faheem}                      & King Saud University, Saudi Arabia                                     & 1                                   \\ 
\textbf{IRAQ}~\cite{aliwy:2020:arabic-dialects}                       & University of Kufa, Iraq                                       & 1                                   \\ \hline
\end{tabular}
~\label{tab:teams}
\caption{List of the 14 teams that participated in Subtasks 1 and 2 \textit{and} submitted description papers.}
\end{table}

\subsection{Shared Task Results}

\begin{table}[t]
\centering
\footnotesize
\begin{tabular}{l|l|l|l|l}
\hline
\multicolumn{1}{c|}{\textbf{Team}} & \multicolumn{1}{c|}{\textbf{F1}} & \multicolumn{1}{c|}{\textbf{Accuracy}} & \multicolumn{1}{c|}{\textbf{Precision}} & \multicolumn{1}{c}{\textbf{Recall}} \\ \hline
\textbf{Mawdoo3 AI}                      & \textbf{26.78 (1)}                       & \textbf{42.86 (1)}                             & \textbf{32.52 (1)}                              & \textbf{25.19 (1)}                           \\ 
\textbf{BERT\_NGRAMS}                    & 25.99 (2)                       & 39.66 (2)                             & 30.26 (2)                              & 24.85 (2)                           \\ 
\textbf{Arabic Processors}               & 23.26 (3)                       & 38.34 (3)                             & 27.17 (4)                              & 22.43 (5)                           \\ 
\textbf{Tri-directional}                 & 23.09 (4)                       & 37.70 (5)                             & 26.40 (5)                              & 23.04 (4)                           \\ 
\textbf{MMZ}                             & 22.58 (5)                       & 38.28 (4)                             & 24.28 (8)                              & 23.36 (3)                           \\ 
\textbf{QMUL Team}                       & 20.77 (6)                       & 34.32 (11)                            & 21.62 (13)                             & 21.09 (6)                           \\ 
\textbf{Code Lyoko}                      & 20.34 (7)                       & 36.26 (8)                             & 27.83 (3)                              & 20.56 (8)                           \\ 
\textbf{TRY\_NLP}                        & 20.04 (8)                       & 33.66 (15)                            & 20.07 (14)                             & 21.07 (7)                           \\ 
\textbf{Sorbonne}                        & 18.80 (9)                       & 36.54 (7)                            & 24.87 (7)                             & 18.05 (12)                           \\ 
\textbf{Iktishaf}                        & 18.63 (10)                      & 33.98 (14)                             & 20.21 (15)                              & 18.76 (9)                          \\ 
\textbf{Speech Translation}              & 18.27 (11)                      & 36.68 (6)                             & 23.75 (10)                             & 18.06 (11)                          \\ 
\textbf{LTG-ST}                          & 17.71 (12)                      & 36.22 (9)                             & 24.93 (6)                              & 17.21 (13)                          \\ 
\textbf{Alexa}                           & 17.29 (13)                      & 34.16 (12)                            & 22.09 (12)                             & 16.81 (15)                          \\ 
\textbf{NAYEL}                           & 16.84 (14)                      & 30.98 (18)                            & 17.88 (16)                             & 18.20 (10)                          \\ 
\textbf{DNLP}                            & 16.50 (15)                      & 31.28 (17)                            & 17.84 (17)                             & 17.04 (14)                          \\ 
\textbf{NLPRL}                           & 15.77 (16)                      & 35.06 (10)                            & 23.96 (9)                              & 15.92 (16)                          \\ 
\textbf{Alpha}                           & 15.10 (17)                      & 34.02 (13)                            & 22.34 (11)                             & 14.71 (17)                          \\ 
\textbf{\color{blue} Our Baseline II}                           & 13.32                      & 32.38                            & 14.57                             & 14.69                          \\

\textbf{IRAQ}                            & 12.45 (18)                      & 31.60 (16)                            & 16.39 (18)                             & 12.67 (18)                          \\
\textbf{\color{blue} Our Baseline I}                            & 1.71                      & 21.84                           & 1.04                             & 4.76                          \\
\hline
\end{tabular}
\caption{Results for Subtask 1. The numbers in parentheses are the ranks. The table is sorted on the $macro-F1$ score, the official metric. Some teams did not submit description papers.}\label{tab:sub1_res}
\end{table}

Table~\ref{tab:sub1_res} presents the best TEST results for all 18 teams who submitted systems for Subtask 1, regardless of whether they have submitted a paper. Based on the official metric, $macro-F_1$, Mawdoo3-AI obtained the best performance with 26.78\% $F_1$ score. Table~\ref{tab:sub2_res} presents the best TEST results of each of the 4 teams who submitted systems to Subtask 2. Team BERT-NGRAMS achieved the best $F_1$ score that is 6.39\%.~\footnote{The full sets of results for Subtask 1 and Subtask 2 are in Tables~\ref{tab:sub1_res_full} and ~\ref{tab:sub2_res_full}, respectively, in Appendix A.}

\begin{table}[t]
\centering
\footnotesize
\begin{tabular}{l|l|l|l|l}
\hline
\multicolumn{1}{c|}{\textbf{Team}} & \multicolumn{1}{c|}{\textbf{F1}} & \multicolumn{1}{c|}{\textbf{Accuracy}} & \multicolumn{1}{c|}{\textbf{Precision}} & \multicolumn{1}{c}{\textbf{Recall}} \\ \hline
\textbf{BERT\_NGRAMS}                    & \textbf{6.39 (1)}                     & 6.50 (2)                             & \textbf{7.84 (1)}                              & 6.54 (2)                           \\ 
\textbf{Arabic Processors}               & 5.75 (2)                       & \textbf{6.80 (1)}                            & 6.78 (2)                              &\textbf{6.74 (1)}                        \\ 
\textbf{NAYEL}                           & 4.99 (3)                       & 5.22 (3)                             & 5.52 (3)                              & 5.17 (3)                           \\ 
\textbf{TRY\_NLP}                        & 4.03 (4)                       & 4.86 (4)                             & 3.74 (4)                              & 4.68 (4)                           \\
\textbf{\color{blue} Our Baseline II}                        & 2.13                       & 3.32                             & 4.04                              & 3.22                          \\

\textbf{\color{blue} Our Baseline I}                        & 0.03                       & 1.92                             & 0.02                              & 1.00                          \\ \hline
\end{tabular}\caption{Results for Subtask 2. The numbers in parentheses are the ranks. Table is sorted on the $macro-F1$ score, the official metric. Team NAYEL does not have a description paper.}\label{tab:sub2_res}
\end{table}

\def\checkmark{\tikz\fill[scale=0.4](0,.35) -- (.25,0) -- (1,.7) -- (.25,.15) -- cycle;} 


\subsection{General Description of Submitted Systems}

In Table~\ref{tab:system_sum}, we provide a high-level description of the systems submitted to each subtask. For each team, we list the overall number of submissions per subtask, their overall best score, the features employed, the methods adopted/developed, and whether they have used the 10M unlabeled tweet dataset we provided to all teams. As can be seen from the table, the majority of the top teams have (1) used Transformers, (2) exploited the unlabeled data for further pre-training, and/or (3) have used self-training to enhance their models. The rest of participating teams have either used a type of neural networks other than Transformers or resorted to linear machine learning models, usually with some form of ensembling.   
\begin{table}[t]
\centering
\footnotesize
\begin{tabular}{lcrccccccccccccc}
\hline
                              & \multicolumn{1}{l}{} & \multicolumn{1}{l}{}   & \multicolumn{5}{c}{\textbf{Features}}                                  & \multicolumn{5}{c}{\textbf{Methods}}                                                                 & \multicolumn{3}{c}{\textbf{Uses Unlabelled 10M}}   \\ \cmidrule(lr){4-8} \cmidrule(lr){9-13} \cmidrule(lr){14-16}
\multicolumn{1}{c}{\textbf{Team }} & {\rotatebox[origin=c]{90}{\textbf{\# Submissions}}}           & {\rotatebox[origin=c]{90}{\textbf{$F_{1}$}}}& {\rotatebox[origin=c]{90}{\textbf{\textit{N}-grams}}} &  {\rotatebox[origin=c]{90}{\textbf{TF-IDF}}}& {\rotatebox[origin=c]{90}{\textbf{Word Embed.}}} & {\rotatebox[origin=c]{90}{\textbf{Topic Models}}} & {\rotatebox[origin=c]{90}{\textbf{Sampling}}}  & {\rotatebox[origin=c]{90}{\textbf{Classical ML}}}  & {\rotatebox[origin=c]{90}{\textbf{Neural Nets}}}  & {\rotatebox[origin=c]{90}{\textbf{Transformer}}}       & {\rotatebox[origin=c]{90}{\textbf{Ensemble}}}                & {\rotatebox[origin=c]{90}{\textbf{Hierarchical}}}  & {\rotatebox[origin=c]{90}{\textbf{Pre-Training}}}  & {\rotatebox[origin=c]{90}{\textbf{Data Augment.}}}  & {\rotatebox[origin=c]{90}{\textbf{Self-Training}}}  \\ \hline
\multicolumn{16}{c}{\textbf{\colorbox{blue!10}{SUBTASK 1}}}                                                                                                                                                                                                                                                                  \\ \hline
\textbf{Mawdoo3 AI}                    & 3                    & 26.78                  & \checkmark      &  \checkmark       &  \checkmark                &                &  \checkmark         &  \checkmark                 &  \checkmark            &  \checkmark                       &  \checkmark                       &             & \checkmark            &              &                 \\
\textbf{BERT\_NGRAMS}                  & 4                    & 25.99                  & \checkmark      & \checkmark      &                 &                &          & \checkmark                &             & \multicolumn{1}{c|}{ \checkmark } & \multicolumn{1}{c|}{ \checkmark } &             &              &  \checkmark             &                 \\ 
\textbf{Arabic Processors}               & 3                    & 23.26                  &  \checkmark       &  \checkmark       &                 &                &          &  \checkmark                 &             &  \checkmark                       &  \checkmark                       &             &              &              &  \checkmark                \\
\textbf{Tri-directional}               & 1                    & 23.09                  &        &        &                 &                &  \checkmark         &  \checkmark                 &             &  \checkmark                       &                        &  \checkmark            &  \checkmark             &              &                 \\
\textbf{MMZ}                            & 2                    & 22.58                  &  \checkmark       &  \checkmark       &                 &                &          &  \checkmark                 &             &  \checkmark                       &  \checkmark                       &             &  \checkmark             &              &                 \\
\textbf{QMUL Team}                        & 2                    & 20.77                  &  \checkmark       &  \checkmark       &  \checkmark                &  \checkmark               &          &                  &  \checkmark            &                        &                        &             &  \checkmark             &              &                 \\
\textbf{Code Lyoko}                     & 2                    & 20.34                  &  \checkmark       &  \checkmark       &  \checkmark                &                &          &  \checkmark                 &  \checkmark            &  \checkmark                       &                        &             &  \checkmark             &              &                 \\
\textbf{TRY\_NLP}                       & 3                    & 20.04                  &  \checkmark       &  \checkmark       &  \checkmark                &                &          &                  &  \checkmark            &  \checkmark                       &                        &             &              &              &  \checkmark                \\
\textbf{Sorbonne}                       & 3                    & 18.80                  &  \checkmark       &  \checkmark       &  \checkmark                &                &          &  \checkmark                 &  \checkmark            &                        &  \checkmark                       &             &  \checkmark             &              &                 \\
\textbf{Speech Translation}             & 3                    & 18.27                  &  \checkmark       &  \checkmark       &                 &                &  \checkmark         &  \checkmark                 &             &                        &  \checkmark                       &             &              &              &                 \\
\textbf{LTG-ST}                        & 2                    & 17.71                  &  \checkmark       &  \checkmark       &                 &                &          &  \checkmark                 &             &                        &  \checkmark                       &             &              &              &                 \\
\textbf{Alexa}                         & 3                    & 17.29                  &  \checkmark       &  \checkmark       &                 &                &          &  \checkmark                 &             &                        &  \checkmark                       &             &              &              &                 \\
\textbf{Alpha}                         & 5                    & 15.10                  &  \checkmark       &  \checkmark       &                 &                &          &  \checkmark                 &             &                        &                        &             &              &              &                 \\
\textbf{IRAQ}                           & 3                    & 12.45                  &  \checkmark       &  \checkmark       &                 &                &          &  \checkmark                 &             &                        &  \checkmark                       &             &              &              &                 \\ \hline
\multicolumn{16}{c}{\textbf{\colorbox{blue!10}{SUBTASK 2}}}                                                                                                                                                                                                                                                                  \\ \hline
\textbf{BERT\_NGRAMS}                   & 3                    & 6.39                   &  \checkmark       &  \checkmark       &                 &                &          &  \checkmark                 &  \checkmark            &  \checkmark                       &  \checkmark                       &  \checkmark            &              &  \checkmark             &                 \\
\textbf{Arabic Processors}              & 1                    & 5.75                   &  \checkmark       &  \checkmark       &                 &                &          &  \checkmark                 &  \checkmark            &  \checkmark                       &                        &             &              &              & \checkmark               \\
\textbf{TRY\_NLP}                     & 2                    & 4.03                   & \checkmark      & \checkmark      & \checkmark               &                &          &                  & \checkmark           & \checkmark                      &                        &             &              &              & \checkmark               \\ \hline
\end{tabular}
\caption{Summary of approaches used by participating teams in Subtasks 1 and 2. Classical ML refers to any non-neural machine learning methods such as naive Bayes and support vector machines. The term ``neural nets" refers to any model based on neural networks (e.g., RNN, CNN) except Transformer models. Transformer refers to neural networks based on a Transformer architecture such as BERT. The table is sorted by official metric , $macro-F_1$. We only list teams that submitted a description paper.}\label{tab:system_sum}
\end{table}
\section{Conclusion and Future Work}\label{sec:conc}
We presented an overview of the NADI 2020 shared task. We described the dataset and identified areas of improvement especially related to its collection. We also provided a high-level description of participating teams. The number of submissions to the shared task reflects an interest in the community and calls for further work in the area of Arabic dialect detection, but also more generally Arabic dialect processing. 

In the future, we plan to host a second iteration of the NADI shared task that will use new datasets and pursue a number of novel questions inspired by the issues discovered in this year's task. For example, in addition to DA classification, we will propose {\it MSA regional use  classification} as a subtask. Since MSA is shared across the Arab world, we hypothesize this will be a more challenging task than DA classification. 
We will also encourage teams to experiment with studying the interaction between MSA and DA in novel ways. For example, questions as to the utility of using DA data to improve MSA
regional use  classification systems and vice versa can be investigated exploiting various machine learning methods.

\section*{Acknowledgments}
We gratefully acknowledge the support of the Natural Sciences and Engineering Research Council of Canada (NSERC), the Social Sciences Research Council of Canada (SSHRC), and Compute Canada (\url{www.computecanada.ca}). We thank AbdelRahim Elmadany for assisting with dataset preparation, setting up the Codalab for the shared task, and providing the map in Figure~\ref{fig:province_map}.

\bibliography{dlnlp,camel-bib-v2,ALLBIB-2.2,NADI-ST-papers.bib,extra}

\begin{thebibliography}{}

\bibitem[\protect\citename{Abbes \bgroup et al.\egroup }2020]{abbes2020daict}
Ines Abbes, Wajdi Zaghouani, Omaima El-Hardlo, and Faten Ashour.
\newblock 2020.
\newblock Daict: A dialectal {Arabic} irony corpus extracted from twitter.
\newblock In {\em Proceedings of The 12th Language Resources and Evaluation
  Conference}, pages 6265--6271.

\bibitem[\protect\citename{Abdul{-}Mageed \bgroup et al.\egroup
  }2018]{Abdul-Mageed:2018:you}
Muhammad Abdul{-}Mageed, Hassan Alhuzali, and Mohamed Elaraby.
\newblock 2018.
\newblock You tweet what you speak: A city-level dataset of {A}rabic dialects.
\newblock In {\em Proceedings of the Language Resources and Evaluation
  Conference (LREC)}, Miyazaki, Japan.

\bibitem[\protect\citename{Abdul-Mageed \bgroup et al.\egroup
  }2020]{mageed2020microdialect}
Muhammad Abdul-Mageed, Chiyu Zhang, AbdelRahim Elmadany, and Lyle Ungar.
\newblock 2020.
\newblock Toward micro-dialect identification in diaglossic and code-switched
  environments.
\newblock {\em arXiv preprint arXiv:2010.04900}.

\bibitem[\protect\citename{Al-Sabbagh and Girju}2012]{al2012yadac}
Rania Al-Sabbagh and Roxana Girju.
\newblock 2012.
\newblock {YADAC: Yet another Dialectal Arabic Corpus}.
\newblock In {\em Proceedings of the Eighth International Conference on
  Language Resources and Evaluation (LREC-2012)}, pages 2882--2889.

\bibitem[\protect\citename{Al-Twairesh \bgroup et al.\egroup
  }2018]{Al-Twairesh:2018:suar}
Nora Al-Twairesh, Rawan Al-Matham, Nora Madi, Nada Almugren, Al-Hanouf
  Al-Aljmi, Shahad Alshalan, Raghad Alshalan, Nafla Alrumayyan, Shams Al-Manea,
  Sumayah Bawazeer, Nourah Al-Mutlaq, Nada Almanea, Waad~Bin Huwaymil, Dalal
  Alqusair, Reem Alotaibi, Suha Al-Senaydi, and Abeer Alfutamani.
\newblock 2018.
\newblock {SUAR}: Towards building a corpus for the {S}audi dialect.
\newblock In {\em Proceedings of the International Conference on {A}rabic
  Computational Linguistics (ACLing)}.

\bibitem[\protect\citename{Aliwy \bgroup et al.\egroup
  }2020]{aliwy:2020:arabic-dialects}
Ahmed Aliwy, Hawraa Taher, and Zena AboAltaheen.
\newblock 2020.
\newblock {Arabic Dialects Identification for All Arabic countries}.
\newblock In {\em {Proceedings of the Fifth {A}rabic Natural Language
  Processing Workshop (WANLP 2020)}}, Barcelona, Spain.

\bibitem[\protect\citename{Aloraini \bgroup et al.\egroup
  }2020]{aloraini:2020:arabic-dialect}
Abdulrahman Aloraini, Ayman Alhelbawy, and Massimo Poesio.
\newblock 2020.
\newblock {The QMUL/HRBDT contribution to the NADI Arabic Dialect
  Identification Shared Task}.
\newblock In {\em {Proceedings of the Fifth {A}rabic Natural Language
  Processing Workshop (WANLP 2020)}}, Barcelona, Spain.

\bibitem[\protect\citename{AlShenaifi and Azmi}2020]{alShenaifi:2020:faheem}
Nouf AlShenaifi and Aqil Azmi.
\newblock 2020.
\newblock {Faheem at NADI shared task: Identifying the dialect of Arabic
  tweet}.
\newblock In {\em {Proceedings of the Fifth {A}rabic Natural Language
  Processing Workshop (WANLP 2020)}}, Barcelona, Spain.

\bibitem[\protect\citename{Althobaiti}2020]{althobaiti2020automatic}
Maha~J Althobaiti.
\newblock 2020.
\newblock Automatic {A}rabic dialect identification systems for written texts:
  {A} survey.
\newblock {\em arXiv preprint arXiv:2009.12622}.

\bibitem[\protect\citename{Badawi}1973]{badawi1973levels}
MS~Badawi.
\newblock 1973.
\newblock {Levels of contemporary Arabic in Egypt}.
\newblock {\em Cairo: D{\^a}r al Ma’{\^a}rif}.

\bibitem[\protect\citename{Balaji and
  Bharathi}2020]{balaji:2020:semi-supervised}
Nitin Balaji and B.~Bharathi.
\newblock 2020.
\newblock { Semi-supervised Fine-grained Approach for Arabic Dialect Detection
  }.
\newblock In {\em {Proceedings of the Fifth {A}rabic Natural Language
  Processing Workshop (WANLP 2020)}}, Barcelona, Spain.

\bibitem[\protect\citename{Beltagy \bgroup et al.\egroup
  }2020]{beltagy:2020:arabic-dialect}
Ahmad Beltagy, Abdelrahman Abouelenin, and Omar ElSherief.
\newblock 2020.
\newblock {Arabic Dialect Identification Using BERT-Based Domain Adaptation}.
\newblock In {\em {Proceedings of the Fifth {A}rabic Natural Language
  Processing Workshop (WANLP 2020)}}, Barcelona, Spain.

\bibitem[\protect\citename{Bni~Younes \bgroup et al.\egroup
  }2020]{younes:2020:team-alexa}
Mutaz Bni~Younes, Nour Al-Khdour, and Mohammad AL-Smadi.
\newblock 2020.
\newblock {Team Alexa at NADI Shared Task}.
\newblock In {\em {Proceedings of the Fifth {A}rabic Natural Language
  Processing Workshop (WANLP 2020)}}, Barcelona, Spain.

\bibitem[\protect\citename{Bouamor \bgroup et al.\egroup
  }2014]{Bouamor:2014:multidialectal}
Houda Bouamor, Nizar Habash, and Kemal Oflazer.
\newblock 2014.
\newblock A multidialectal parallel corpus of {A}rabic.
\newblock In {\em Proceedings of the Language Resources and Evaluation
  Conference (LREC)}, Reykjavik, Iceland.

\bibitem[\protect\citename{Bouamor \bgroup et al.\egroup
  }2018]{Bouamor:2018:madar}
Houda Bouamor, Nizar Habash, Mohammad Salameh, Wajdi Zaghouani, Owen Rambow,
  Dana Abdulrahim, Ossama Obeid, Salam Khalifa, Fadhl Eryani, Alexander
  Erdmann, and Kemal Oflazer.
\newblock 2018.
\newblock {The MADAR {A}rabic Dialect Corpus and Lexicon}.
\newblock In {\em Proceedings of the Language Resources and Evaluation
  Conference (LREC)}, Miyazaki, Japan.

\bibitem[\protect\citename{Bouamor \bgroup et al.\egroup
  }2019]{bouamor2019madar}
Houda Bouamor, Sabit Hassan, and Nizar Habash.
\newblock 2019.
\newblock {The MADAR shared task on Arabic fine{-}grained dialect
  identification}.
\newblock In {\em {Proceedings of the Fourth Arabic Natural Language Processing
  Workshop}}, pages 199--207.

\bibitem[\protect\citename{Brustad}2000]{Brustad:2000:syntax}
Kristen Brustad.
\newblock 2000.
\newblock {\em The Syntax of Spoken {A}rabic: A Comparative Study of Moroccan,
  Egyptian, Syrian, and Kuwaiti Dialects}.
\newblock Georgetown University Press.

\bibitem[\protect\citename{Cowell}1964]{Cowell:1964:reference}
Mark~W. Cowell.
\newblock 1964.
\newblock {\em {A Reference Grammar of Syrian {A}rabic}}.
\newblock Georgetown University Press, Washington, D.C.

\bibitem[\protect\citename{Diab \bgroup et al.\egroup }2010]{diab2010colaba}
Mona Diab, Nizar Habash, Owen Rambow, Mohamed Altantawy, and Yassine Benajiba.
\newblock 2010.
\newblock {COLABA: Arabic dialect annotation and processing}.
\newblock In {\em {LREC workshop on Semitic language processing}}, pages
  66--74.

\bibitem[\protect\citename{El-Haj}2020]{el-haj-2020-habibi}
Mahmoud El-Haj.
\newblock 2020.
\newblock Habibi - a multi dialect multi national {A}rabic song lyrics corpus.
\newblock In {\em Proceedings of the 12th Language Resources and Evaluation
  Conference}, pages 1318--1326, Marseille, France, May.

\bibitem[\protect\citename{El~Mekki \bgroup et al.\egroup
  }2020]{mekki:2020:weighted}
Abdellah El~Mekki, Ahmed Alami, Hamza Alami, Ahmed Khoumsi, and Ismail Berrada.
\newblock 2020.
\newblock {Weighted combination of BERT and N-GRAM features for Nuanced Arabic
  Dialect Identification}.
\newblock In {\em {Proceedings of the Fifth {A}rabic Natural Language
  Processing Workshop (WANLP 2020)}}, Barcelona, Spain.

\bibitem[\protect\citename{Elfardy \bgroup et al.\egroup
  }2014]{Elfardy:2014:aida}
Heba Elfardy, Mohamed Al-Badrashiny, and Mona Diab.
\newblock 2014.
\newblock Aida: Identifying code switching in informal {A}rabic text.
\newblock In {\em Proceedings of the Conference on Empirical Methods in Natural
  Language Processing (EMNLP)}, pages 94--101, Doha, Qatar.

\bibitem[\protect\citename{Gaanoun and
  Benelallam}2020]{gaanoun:2020:arabicprocessors}
Kamel Gaanoun and Imade Benelallam.
\newblock 2020.
\newblock {Arabic dialect identification: An Arabic-BERT model with data
  augmentation and ensembling strategy}.
\newblock In {\em {Proceedings of the Fifth {A}rabic Natural Language
  Processing Workshop (WANLP 2020)}}, Barcelona, Spain.

\bibitem[\protect\citename{Gadalla \bgroup et al.\egroup
  }1997]{Gadalla:1997:callhome}
Hassan Gadalla, Hanaa Kilany, Howaida Arram, Ashraf Yacoub, Alaa El-Habashi,
  Amr Shalaby, Krisjanis Karins, Everett Rowson, Robert MacIntyre, Paul
  Kingsbury, David Graff, and Cynthia McLemore.
\newblock 1997.
\newblock {CALLHOME} {E}gyptian {A}rabic transcripts {LDC97T19}.
\newblock Web Download. Philadelphia: Linguistic Data Consortium.

\bibitem[\protect\citename{Ghoul and Lejeune}2020]{ghoul:2020:comparison}
Dhaou Ghoul and Gael Lejeune.
\newblock 2020.
\newblock {Comparison between Voting Classifier and Deep Learning methods for
  Arabic Dialect Identification}.
\newblock In {\em {Proceedings of the Fifth {A}rabic Natural Language
  Processing Workshop (WANLP 2020)}}, Barcelona, Spain.

\bibitem[\protect\citename{Harrell}1962]{Harrell:1962:short}
R.S. Harrell.
\newblock 1962.
\newblock {\em A Short Reference Grammar of Moroccan {A}rabic: With Audio CD}.
\newblock Georgetown classics in {A}rabic language and linguistics. Georgetown
  University Press.

\bibitem[\protect\citename{Holes}2004]{Holes:2004:modern}
Clive Holes.
\newblock 2004.
\newblock {\em {Modern {A}rabic: Structures, Functions, and Varieties}}.
\newblock Georgetown Classics in {A}rabic Language and Linguistics. Georgetown
  University Press.

\bibitem[\protect\citename{Jarrar \bgroup et al.\egroup
  }2016]{Jarrar:2016:curras}
Mustafa Jarrar, Nizar Habash, Faeq Alrimawi, Diyam Akra, and Nasser Zalmout.
\newblock 2016.
\newblock {Curras: an annotated corpus for the Palestinian {A}rabic dialect}.
\newblock {\em Language Resources and Evaluation}, pages 1--31.

\bibitem[\protect\citename{Khalifa \bgroup et al.\egroup
  }2016]{Khalifa:2016:large}
Salam Khalifa, Nizar Habash, Dana Abdulrahim, and Sara Hassan.
\newblock 2016.
\newblock {A Large Scale Corpus of Gulf {A}rabic}.
\newblock In {\em Proceedings of the Language Resources and Evaluation
  Conference (LREC)}, Portoro\v{z}, Slovenia.

\bibitem[\protect\citename{Lichouri and Abbas}2020]{lichouri:2020:speechtrans}
Mohamed Lichouri and Mourad Abbas.
\newblock 2020.
\newblock {Simple vs Oversampling-based Classification Methods for Fine Grained
  Arabic Dialect Identification in Twitter}.
\newblock In {\em {Proceedings of the Fifth {A}rabic Natural Language
  Processing Workshop (WANLP 2020)}}, Barcelona, Spain.

\bibitem[\protect\citename{Malmasi \bgroup et al.\egroup
  }2016]{malmasi2016discriminating}
Shervin Malmasi, Marcos Zampieri, Nikola Ljube{\v{s}}i{\'c}, Preslav Nakov,
  Ahmed Ali, and J{\"o}rg Tiedemann.
\newblock 2016.
\newblock Discriminating between similar languages and arabic dialect
  identification: A report on the third dsl shared task.
\newblock In {\em Proceedings of the third workshop on NLP for similar
  languages, varieties and dialects (VarDial3)}, pages 1--14.

\bibitem[\protect\citename{Mansour \bgroup et al.\egroup
  }2020]{mansour:2020:arabic-dialect}
Moataz Mansour, Moustafa Tohamy, Zeyad Ezzat, and Marwan Torki.
\newblock 2020.
\newblock {Arabic Dialect Identification Using BERT Fine-Tuning}.
\newblock In {\em {Proceedings of the Fifth {A}rabic Natural Language
  Processing Workshop (WANLP 2020)}}, Barcelona, Spain.

\bibitem[\protect\citename{Meftouh \bgroup et al.\egroup
  }2015]{Meftouh:2015:machine}
Karima Meftouh, Salima Harrat, Salma Jamoussi, Mourad Abbas, and Kamel Smaili.
\newblock 2015.
\newblock Machine translation experiments on padic: A parallel {A}rabic dialect
  corpus.
\newblock In {\em Proceedings of the Pacific Asia Conference on Language,
  Information and Computation}.

\bibitem[\protect\citename{Mubarak and Darwish}2014]{Mubarak:2014:using}
Hamdy Mubarak and Kareem Darwish.
\newblock 2014.
\newblock Using {T}witter to collect a multi-dialectal corpus of {A}rabic.
\newblock In {\em Proceedings of the Workshop for {A}rabic Natural Language
  Processing (WANLP)}, Doha, Qatar.

\bibitem[\protect\citename{Mubarak \bgroup et al.\egroup
  }2020]{mubarak2020overview}
Hamdy Mubarak, Kareem Darwish, Walid Magdy, Tamer Elsayed, and Hend Al-Khalifa.
\newblock 2020.
\newblock Overview of osact4 {Arabic} offensive language detection shared task.
\newblock In {\em Proceedings of the 4th Workshop on Open-Source {Arabic}
  Corpora and Processing Tools, with a Shared Task on Offensive Language
  Detection}, pages 48--52.

\bibitem[\protect\citename{Obeid \bgroup et al.\egroup
  }2019]{obeid-etal-2019-adida}
Ossama Obeid, Mohammad Salameh, Houda Bouamor, and Nizar Habash.
\newblock 2019.
\newblock {ADIDA}: Automatic dialect identification for {A}rabic.
\newblock In {\em Proceedings of the 2019 Conference of the North {A}merican
  Chapter of the Association for Computational Linguistics (Demonstrations)},
  pages 6--11, Minneapolis, Minnesota, June. Association for Computational
  Linguistics.

\bibitem[\protect\citename{Sadat \bgroup et al.\egroup
  }2014]{sadat2014automatic}
Fatiha Sadat, Farnazeh Kazemi, and Atefeh Farzindar.
\newblock 2014.
\newblock {Automatic identification of Arabic language varieties and dialects
  in social media}.
\newblock {\em {Proceedings of SocialNLP}}, page~22.

\bibitem[\protect\citename{Salameh \bgroup et al.\egroup
  }2018]{Salameh:2018:fine-grained}
Mohammad Salameh, Houda Bouamor, and Nizar Habash.
\newblock 2018.
\newblock Fine-grained {A}rabic dialect identification.
\newblock In {\em Proceedings of the International Conference on Computational
  Linguistics (COLING)}, pages 1332--1344, Santa Fe, New Mexico, USA.

\bibitem[\protect\citename{Sma{\"\i}li \bgroup et al.\egroup
  }2014]{Smaili:2014:building}
Kamel Sma{\"\i}li, Mourad Abbas, Karima Meftouh, and Salima Harrat.
\newblock 2014.
\newblock Building resources for {Algerian} {{A}rabic} dialects.
\newblock In {\em Proceedings of the Conference of the International Speech
  Communication Association (Interspeech)}.

\bibitem[\protect\citename{Tahssin \bgroup et al.\egroup
  }2020]{tahssin:2020:identifying}
Rawan Tahssin, Youssef Kishk, and Marwan Torki.
\newblock 2020.
\newblock {Identifying Nuanced Dialect for Arabic Tweets with Deep Learning and
  Reverse Translation Corpus Extension System}.
\newblock In {\em {Proceedings of the Fifth {A}rabic Natural Language
  Processing Workshop (WANLP 2020)}}, Barcelona, Spain.

\bibitem[\protect\citename{Talafha \bgroup et al.\egroup
  }2020]{talafha:2020:multi-dialect}
Bashar Talafha, Mohamed Ali, Muhy~Eddin Za’ter, Haitham Seelawi, Ibraheem
  Tuffaha, Mostafa Samir, Wael Farhan, and Hussein~T. AL-NATSHEH.
\newblock 2020.
\newblock {Multi-dialect Arabic BERT for Country-level Dialect Identification}.
\newblock In {\em {Proceedings of the Fifth {A}rabic Natural Language
  Processing Workshop (WANLP 2020)}}, Barcelona, Spain.

\bibitem[\protect\citename{Touileb}2020]{touileb:2020:ltg-st}
Samia Touileb.
\newblock 2020.
\newblock {LTG-ST at NADI Shared Task 1: Arabic Dialect Identification using a
  Stacking Classifier}.
\newblock In {\em {Proceedings of the Fifth {A}rabic Natural Language
  Processing Workshop (WANLP 2020)}}, Barcelona, Spain.

\bibitem[\protect\citename{Zaghouani and Charfi}2018]{Zaghouani:2018:araptweet}
Wajdi Zaghouani and Anis Charfi.
\newblock 2018.
\newblock {ArapTweet: A Large Multi-Dialect Twitter Corpus for Gender, Age and
  Language Variety Identification}.
\newblock In {\em Proceedings of the Language Resources and Evaluation
  Conference (LREC)}, Miyazaki, Japan.

\bibitem[\protect\citename{Zaidan and Callison-Burch}2011]{zaidan2011arabic}
Omar~F Zaidan and Chris Callison-Burch.
\newblock 2011.
\newblock {The Arabic online commentary dataset: an annotated dataset of
  informal Arabic with high dialectal content}.
\newblock In {\em {Proceedings of the 49th Annual Meeting of the Association
  for Computational Linguistics: Human Language Technologies: short
  papers-Volume 2}, Organization = {Association for Computational
  Linguistics}}, pages 37--41.

\bibitem[\protect\citename{Zampieri \bgroup et al.\egroup
  }2017]{zampieri2017findings}
Marcos Zampieri, Shervin Malmasi, Nikola Ljube{\v{s}}i{\'c}, Preslav Nakov,
  Ahmed Ali, J{\"o}rg Tiedemann, Yves Scherrer, and No{\"e}mi Aepli.
\newblock 2017.
\newblock Findings of the vardial evaluation campaign 2017.

\bibitem[\protect\citename{Zampieri \bgroup et al.\egroup
  }2018]{zampieri2018language}
Marcos Zampieri, Shervin Malmasi, Preslav Nakov, Ahmed Ali, Suwon Shon, James
  Glass, Yves Scherrer, Tanja Samardzic, Nikola Ljube{\v{s}}i{\'c}, J{\"o}rg
  Tiedemann, et~al.
\newblock 2018.
\newblock Language identification and morphosyntactic tagging: The second
  vardial evaluation campaign.
\newblock In {\em Proceedings of the Fifth Workshop on NLP for Similar
  Languages, Varieties and Dialects (VarDial 2018)}, pages 1--17.

\end{thebibliography}
\bibliographystyle{acl}
\newpage
\appendix
\appendixpage
\addappheadtotoc
\setcounter{table}{0}  \renewcommand{\thetable}{\Alph{section}\arabic{table}}
\section{Appendix}\label{sec:appendix}
\begin{table}[h!]
\centering
\footnotesize 
\begin{tabular}{l|r|r|r|l|r|r|r}
\hline
\multicolumn{1}{c|}{\multirow{2}{*}{\textbf{Province Name}}} & \multicolumn{3}{c|}{\textbf{\# of Tweets}}                                                  & \multicolumn{1}{c|}{\multirow{2}{*}{\textbf{Province Name}}} & \multicolumn{3}{c}{\textbf{\# of Tweets}}                                                   \\ \cline{2-4} \cline{6-8} 
\multicolumn{1}{c|}{}                            & \multicolumn{1}{c|}{\textbf{train}} & \multicolumn{1}{c|}{\textbf{dev}} & \multicolumn{1}{c|}{\textbf{test}} & \multicolumn{1}{c|}{}                            & \multicolumn{1}{c|}{\textbf{train}} & \multicolumn{1}{c|}{\textbf{dev}} & \multicolumn{1}{c}{\textbf{test}} \\ \hline
Damascus City                                     & 214                        & 53                       & 52                        & Minya                                            & 213                        & 53                       & 52                        \\ 
Ariana                                            & 212                        & 52                       & 52                        & BBordj Bou Arreridj                            & 213                        & 53                       & 52                        \\ 
Asyut                                             & 213                        & 53                       & 52                        & Hawalli                                          & 210                        & 51                       & 51                        \\ 
Marrakech-Tensift-Al Haouz                        & 214                        & 53                       & 52                        & Al Butnan                                        & 214                        & 53                       & 53                        \\ 
Bouira                                            & 213                        & 53                       & 52                        & Abu Dhabi                                        & 214                        & 53                       & 52                        \\ 
Ash Sharqiyah                                     & 32                         & 10                       & 8                         & Kairouan                                         & 114                        & 8                        & 52                        \\ 
Khenchela                                         & 213                        & 53                       & 52                        & Banaadir                                         & 210                        & 51                       & 51                        \\ 
As-Sulaymaniyah                                   & 213                        & 53                       & 52                        & Ar Riyad                                         & 213                        & 54                       & 52                        \\ 
Ad Dakhiliyah                                     & 213                        & 53                       & 52                        & Baghdad                                          & 213                        & 53                       & 52                        \\ 
Fujairah                                          & 214                        & 53                       & 52                        & Djibouti                                         & 210                        & 10                       & 51                        \\ 
An-Najaf                                          & 213                        & 53                       & 52                        & Musandam                                         & 213                        & 27                       & 52                        \\ 
Oriental                                          & 214                        & 37                       & 52                        & Muscat                                           & 213                        & 53                       & 52                        \\ 
Ibb                                               & 213                        & 50                       & 52                        & Doha                                             & 24                         & 52                       & 10                        \\ 
Al Quassim                                        & 213                        & 53                       & 52                        & Khartoum                                         & 210                        & 51                       & 51                        \\ 
Qena                                              & 213                        & 53                       & 52                        & Aswan                                            & 213                        & 53                       & 52                        \\ 
Sousse                                            & 212                        & 52                       & 52                        & North Sinai                                      & 213                        & 53                       & 52                        \\ 
As-Suwayda                                        & 214                        & 53                       & 52                        & Ash Sharqiyah                                    & 395                        & 97                       & 96                        \\ 
Ha'il                                             & 213                        & 54                       & 52                        & Beni Suef                                        & 213                        & 53                       & 52                        \\ 
Jizan                                             & 213                        & 53                       & 52                        & Tabuk                                            & 213                        & 53                       & 52                        \\ 
Jijel                                             & 213                        & 53                       & 52                        & Tripoli                                          & 214                        & 53                       & 53                        \\ 
Mahdia                                            & 212                        & 52                       & 52                        & Béchar                                          & 213                        & 41                       & 52                        \\ 
Ismailia                                          & 213                        & 53                       & 52                        & Najran                                           & 213                        & 53                       & 52                        \\ 
Meknes-Tafilalet                                  & 214                        & 53                       & 52                        & West Bank                                        & 210                        & 51                       & 51                        \\ 
Wasit                                             & 213                        & 53                       & 52                        & Alexandria                                       & 213                        & 53                       & 52                        \\ 
Gaza Strip                                        & 210                        & 51                       & 51                        & Dhofar                                           & 213                        & 53                       & 52                        \\ 
Kafr el-Sheikh                                    & 213                        & 53                       & 52                        & Capital                                          & 210                        & 8                        & 20                        \\ 
Nouakchott                                        & 210                        & 40                       & 5                         & Misrata                                          & 214                        & 53                       & 53                        \\ 
Gharbia                                           & 213                        & 53                       & 52                        & Aqaba                                            & 213                        & 52                       & 52                        \\ 
Al Anbar                                          & 213                        & 53                       & 52                        & Cairo                                            & 213                        & 10                       & 52                        \\ 
Arbil                                             & 213                        & 53                       & 52                        & North Lebanon                                    & 213                        & 52                       & 52                        \\ 
Akkar                                             & 213                        & 6                        & 52                        & South Lebanon                                    & 213                        & 52                       & 52                        \\ 
Makkah                                            & 213                        & 54                       & 52                        & Faiyum                                           & 213                        & 53                       & 52                        \\ 
Hims                                              & 214                        & 53                       & 52                        & Souss-Massa-Draa                                 & 214                        & 53                       & 52                        \\ 
Benghazi                                          & 214                        & 53                       & 53                        & Beheira                                          & 213                        & 53                       & 52                        \\ 
Lattakia                                          & 214                        & 53                       & 52                        & Al Jabal al Akhdar                               & 214                        & 53                       & 53                        \\ 
Port Said                                         & 213                        & 53                       & 52                        & Ouargla                                          & 213                        & 53                       & 52                        \\ 
Oran                                              & 213                        & 53                       & 52                        & Monufia                                          & 213                        & 53                       & 52                        \\ 
Aden                                              & 213                        & 52                       & 52                        & Sohag                                            & 213                        & 53                       & 52                        \\ 
Al Madinah                                        & 213                        & 54                       & 52                        & Al Batnah                                        & 214                        & 53                       & 52                        \\ 
Red Sea                                           & 213                        & 53                       & 52                        & Dubai                                            & 214   & 53 & 52  \\ 
Karbala                                           & 213                        & 53                       & 52                        & Maysan                                           & 213   & 53 & 52  \\ 
Zarqa                                             & 213                        & 52                       & 52                        & Ninawa                                           & 213   & 53 & 52  \\ 
Basra                                             & 213                        & 53                       & 52                        & Al Hudaydah                                      & 213   & 52 & 52  \\ 
Suez                                              & 213                        & 53                       & 52                        & Jahra                                            & 210   & 19 & 51  \\ 
Dakahlia                                          & 213                        & 53                       & 52                        & Al-Muthannia                                     & 213   & 53 & 52  \\ 
South Sinai                                       & 213                        & 53                       & 52                        & Ras Al Khaymah                                   & 214   & 53 & 52  \\ 
Umm Al Qaywayn                                    & 214                        & 53                       & 5                         & Dihok                                            & 213   & 53 & 52  \\ 
Aleppo                                            & 214                        & 53                       & 52                        & Ar Rayyan                                        & 210   & 52 & 51  \\ 
Tanger-Tetouan                                    & 214                        & 53                       & 52                        & Luxor                                            & 213   & 53 & 52  \\ 
Asir                                              & 213                        & 54                       & 52                        & Dhamar                                           & 212   & 52 & 23  \\ \hline
\end{tabular}
\caption{Distribution of the NADI data over provinces, by country, across our TRAIN, DEV, and TEST splits.}\label{tab:data_provinces} 
\end{table}

\begin{table}[h]
\centering
\footnotesize
\begin{tabular}{lllll}
\hline
\multicolumn{1}{c}{\textbf{Team Name}} & \multicolumn{1}{c}{\textbf{F1}} & \multicolumn{1}{c}{\textbf{Accuracy}} & \multicolumn{1}{c}{\textbf{Precision}} & \multicolumn{1}{c}{\textbf{Recall}} \\ \hline
Mawdoo3 AI                             & \textbf{26.78 (1)}                      & 42.86 (2)                            & \textbf{32.52 (1)}                             & 25.19 (2)                          \\
Mawdoo3 AI                             & 26.77 (2)                      & 42.56 (3)                            & 31.51 (4)                             & \textbf{25.45 (1)}                          \\
Mawdoo3 AI                             & 26.47 (3)                      & \textbf{43.18 (1)}                            & 31.59 (3)                             & 25.12 (3)                          \\
BERT\_NGRAMS                           & 25.99 (4)                      & 39.66 (5)                            & 30.26 (6)                             & 24.85 (4)                          \\
BERT\_NGRAMS                           & 25.99 (4)                      & 39.66 (5)                            & 30.26 (6)                             & 24.85 (4)                          \\
BERT\_NGRAMS                           & 25.02 (5)                      & 38.92 (6)                            & 30.92 (5)                             & 23.81 (5)                          \\
BERT\_NGRAMS                           & 23.83 (6)                      & 40.88 (4)                            & 32.50 (2)                             & 23.36 (7)                          \\
Arabic Processors                       & 23.26 (7)                      & 38.34 (8)                            & 27.17 (9)                             & 22.43 (9)                          \\
Tri-directional                        & 23.09 (8)                      & 37.70 (10)                           & 26.40 (11)                            & 23.04 (8)                          \\
Arabic Processors                       & 23.03 (9)                      & 38.42 (7)                            & 27.40 (8)                             & 22.40 (10)                         \\
MMZ                                    & 22.58 (10)                     & 38.28 (9)                            & 24.28 (15)                            & 23.36 (6)                          \\
MMZ                                    & 22.58 (10)                     & 38.28 (9)                            & 24.28 (15)                            & 23.36 (6)                          \\
Arabic Processors                       & 22.52 (11)                     & 38.28 (9)                            & 26.70 (10)                            & 22.12 (11)                         \\
QMUL team                               & 20.77 (12)                     & 34.32 (22)                           & 21.62 (28)                            & 21.09 (12)                         \\
Code Lyoko                              & 20.34 (13)                     & 36.26 (13)                           & 27.83 (7)                             & 20.56 (15)                         \\
TRY\_NLP                               & 20.04 (14)                     & 33.66 (27)                           & 20.70 (29)                            & 21.07 (13)                         \\
TRY\_NLP                               & 20.01 (15)                     & 33.58 (28)                           & 20.66 (30)                            & 21.03 (14)                         \\
TRY\_NLP                               & 19.84 (16)                     & 34.80 (20)                           & 20.54 (31)                            & 20.17 (16)                         \\
QMUL team                               & 19.45 (17)                     & 33.74 (26)                           & 20.40 (33)                            & 19.84 (17)                         \\
Sorbonne                               & 18.80 (18)                     & 36.54 (12)                           & 24.87 (14)                            & 18.05 (21)                         \\
Iktishaf                               & 18.63 (19)                     & 33.98 (25)                           & 20.21 (34)                            & 18.76 (18)                         \\
Speech Translation                      & 18.27 (20)                     & 36.68 (11)                           & 23.75 (20)                            & 18.06 (20)                         \\
Speech Translation                      & 17.90 (21)                     & 35.68 (16)                           & 22.40 (23)                            & 17.64 (23)                         \\
Iktishaf                               & 17.84 (22)                     & 33.48 (29)                           & 19.07 (36)                            & 17.98 (22)                         \\
Sorbonne                               & 17.77 (23)                     & 35.44 (18)                           & 23.79 (19)                            & 17.15 (26)                         \\
LTG-ST                                 & 17.71 (24)                     & 36.22 (15)                           & 24.93 (13)                            & 17.21 (25)                         \\
Speech Translation                      & 17.69 (25)                     & 36.24 (14)                           & 22.17 (25)                            & 17.41 (24)                         \\
Alexa                                  & 17.29 (26)                     & 34.16 (23)                           & 22.09 (26)                            & 16.81 (29)                         \\
Alexa                                  & 17.20 (27)                     & 35.64 (17)                           & 23.53 (21)                            & 16.86 (28)                         \\
NAYEL                                  & 16.84 (28)                     & 30.98 (41)                           & 17.88 (38)                            & 18.20 (19)                         \\
LTG-ST                                 & 16.81 (29)                     & 34.78 (21)                           & 23.90 (18)                            & 16.46 (31)                         \\
DNLP                                   & 16.50 (30)                     & 31.28 (39)                           & 17.84 (39)                            & 17.04 (27)                         \\
DNLP                                   & 16.27 (31)                     & 31.24 (40)                           & 17.66 (40)                            & 16.77 (30)                         \\
Sorbonne                               & 16.06 (32)                     & 31.90 (36)                           & 22.00 (27)                            & 15.90 (34)                         \\
NAYEL                                  & 15.81 (33)                     & 32.22 (35)                           & 17.91 (37)                            & 16.01 (32)                         \\
NLPRL                                  & 15.77 (34)                     & 35.06 (19)                           & 23.96 (17)                            & 15.92 (33)                         \\
Alpha                                  & 15.10 (35)                     & 34.02 (24)                           & 22.34 (24)                            & 14.71 (39)                         \\
Alexa                                  & 15.09 (36)                     & 34.78 (21)                           & 23.00 (22)                            & 15.68 (35)                         \\
Alpha                                  & 14.91 (37)                     & 32.80 (32)                           & 25.43 (12)                            & 14.30 (42)                         \\
Alpha                                  & 14.72 (38)                     & 33.00 (30)                           & 20.44 (32)                            & 14.61 (40)                         \\
Alpha                                  & 14.61 (39)                     & 32.24 (34)                           & 17.30 (41)                            & 14.81 (37)                         \\
NAYEL                                  & 14.37 (40)                     & 29.42 (42)                           & 15.32 (45)                            & 14.90 (36)                         \\
Alpha                                  & 14.27 (41)                     & 32.84 (31)                           & 24.15 (16)                            & 14.33 (41)                         \\
Sorbonne                               & 14.21 (42)                     & 32.38 (33)                           & 19.13 (35)                            & 14.25 (43)                         \\
Code Lyoko                              & 13.57 (43)                     & 31.70 (37)                           & 15.32 (44)                            & 14.74 (38)                         \\
IRAQ                                   & 12.45 (44)                     & 31.60 (38)                           & 16.39 (42)                            & 12.67 (44)                         \\
IRAQ                                   & 12.20 (45)                     & 31.28 (39)                           & 15.76 (43)                            & 12.45 (45)                         \\ \hline
\end{tabular}
\caption{Full results for Subtask 1. The numbers in parentheses are the ranks. The table is sorted on the  $macro ~F_1$ score, the official metric.}\label{tab:sub1_res_full} 
\end{table}

\begin{table}[h]
\footnotesize
\centering
\begin{tabular}{l|r|r|r|r}
\hline
\multicolumn{1}{c|}{\textbf{Team Name}} & \multicolumn{1}{c|}{\textbf{F1}} & \multicolumn{1}{c|}{\textbf{Accuracy}} & \multicolumn{1}{c|}{\textbf{Precision}} & \multicolumn{1}{c}{\textbf{Recall}} \\ \hline
BERT\_NGRAMS                   & \textbf{6.39 (1)}                       & 6.50 (2)                             & 7.84 (2)                              & 6.54 (2)                           \\ 
BERT\_NGRAMS                  & 6.08 (2)                       & 6.16 (3)                             & 7.78 (3)                              & 6.03 (3)                           \\ 
Arabic Processors                & 5.75 (3)                       &\textbf{6.80 (1)}                           & 6.78 (4)                              & \textbf{6.74 (1)}                          \\ 
BERT\_NGRAMS                   & 5.42 (4)                       & 5.32 (4)                             & \textbf{8.00 (1)}                              & 5.24 (4)                           \\ 
NAYEL                           & 4.99 (5)                       & 5.22 (5)                             & 5.52 (5)                              & 5.17 (5)                           \\ 
NAYEL                           & 4.28 (6)                       & 4.48 (8)                             & 4.34 (6)                              & 4.69 (6)                           \\ 
TRY\_NLP                       & 4.03 (7)                       & 4.86 (6)                             & 3.74 (9)                              & 4.68 (7)                           \\ 
TRY\_NLP                        & 3.94 (8)                       & 4.54 (7)                             & 3.86 (7)                              & 4.45 (8)                           \\ 
NAYEL                           & 3.60 (9)                       & 3.84 (9)                             & 3.83 (8)                              & 3.85 (9)                           \\ \hline
\end{tabular}
\caption{Full results for Subtask 2. The numbers in parentheses are the ranks. The table is sorted on the  $macro ~F_1$ score, the official metric.}\label{tab:sub2_res_full}
\end{table}

\end{document}